\documentclass[letterpaper, 10 pt, conference]{ieeeconf}  % Comment this line out if you need a4paper

\IEEEoverridecommandlockouts                              % This command is only needed if 
\usepackage{graphics} % for pdf, bitmapped graphics files
\usepackage{epsfig} % for postscript graphics files
\usepackage{mathptmx} % assumes new font selection scheme installed
\usepackage{times} % assumes new font selection scheme installed
\usepackage{amsmath} % assumes amsmath package installed
\usepackage{amssymb}  % assumes amsmath package installed
 \usepackage[hidelinks]{hyperref}
\usepackage{arydshln}
\usepackage[font=small]{caption}

\title{\LARGE \bf
Multi-Modal Depth Estimation Using Convolutional Neural Networks
}

\author{Sadique Adnan Siddiqui, Axel Vierling and Karsten Berns\\
Robotics Research Lab, Dep. of Computer Science, TU Kaiserslautern, Kaiserslautern, Germany% <-this % stops a space
}

\begin{document}

\maketitle
\thispagestyle{empty}
\pagestyle{empty}

%%%%%%%%%%%%%%%%%%%%%%%%%%%%%%%%%%%%%%%%%%%%%%%%%%%%%%%%%%%%%%%%%%%%%%%%%%%%%%%%
\begin{abstract}
This paper addresses the problem of dense depth predictions from sparse distance sensor data and a single camera image on challenging weather conditions. This work explores the significance of different sensor modalities such as camera, Radar, and Lidar for estimating depth by applying Deep Learning approaches. Although Lidar has higher depth-sensing abilities than Radar and has been integrated with camera images in lots of previous works, depth estimation using CNN's on the fusion of robust Radar distance data and camera images has not been explored much. In this work, a deep regression network is proposed utilizing a transfer learning approach consisting of an encoder where a high performing pre-trained model has been used to initialize it for extracting dense features and a decoder for upsampling and predicting desired depth. The results are demonstrated on Nuscenes, KITTI, and a Synthetic dataset which was created using the CARLA simulator. Also, top-view zoom-camera images captured from the crane on a construction site are evaluated to estimate the distance of the crane boom carrying heavy loads from the ground to show the usability in safety-critical applications. 
\end{abstract}

%%%%%%%%%%%%%%%%%%%%%%%%%%%%%%%%%%%%%%%%%%%%%%%%%%%%%%%%%%%%%%%%%%%%%%%%%%%%%%%%
\section{INTRODUCTION}
In recent years, there has been a meteoric rise in autonomous driving and automation research paired with an undeniable success of machine learning approaches on computer vision tasks. Although advanced driving assistance systems could significantly reduce the accidents caused by human-controlled driving, there is a lot of room for improvement concerning human's safety. Fully autonomous vehicles being driven in every part of the world is a distant dream, achieving a standard level of autonomy with the help of sustainable assistance systems that could help in mitigating the chances of crashes due to human error is the pressing priority.   %Maybe focus more on safety systems/assistance systems than autonomous driving
Sense of depth is crucial for understanding the world around us. Humans can easily assess the distance between the objects through their binocular vision. Intelligent systems do not have insights that a human has, they need to assign specific features for every object visible on the scene so that they could recognize them and infer their semantics and geometric structure. For computer vision in the autonomous driving field, the sensors continuously capture the information either in the form of images or point clouds to build a representation that could be understandable to the vehicle. Depth information i.e distance information of the objects from the vehicle captured from depth sensors is vital for dynamic driving activities such as avoiding collisions from the obstacles. Several sensors are used to estimate robust depth such as stereo cameras, depth cameras, Lidars but they have their limitations. Lidars are usually expensive and result in sparse depth point clouds, whereas stereo cameras require careful calibration of both the cameras. Stereo cameras often result in mediocre performance on low light, object occluded and less textured regions. Motion sensors such as the Kinect are sensitive to sunlight and are not suitable for outdoor applications.
\begin{figure}[t]
\centering
\includegraphics[width=0.45\textwidth]{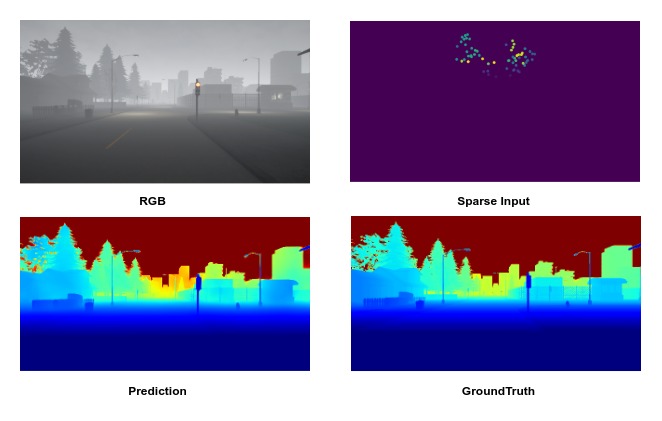}
\caption{Prediction from deep regression network trained by fusing camera image on different weather conditions (such as fog as depicted in the figure) and sparse depth input of Radar.}
\label{Early-Fusion}
\end{figure}
\raggedbottom
% Is this needed? very general
Due to these limitations depth estimation using Deep Learning approaches gained much attention in recent years and their fusion with a robust depth sensor such as a Radar seems promising.
The significant contribution of this work is a Convolutional Neural Network based regression network that takes fused input of a sparse point cloud data from Radar and camera image and it predicts a full-resolution depth image for images recorded on challenging weather conditions such as day, night, cloudy, fog, and rain. The positive aspect of using Radars is its robustness to adverse weather conditions. A Radar sensor is cheap, lightweight, small in size, and is capable of measuring a more considerable distance than Lidars. RGB images can provide information about the appearance and texture of the scene, whereas depth sensors provide information about the shape of the object, and it is invariant to color alteration and illumination \cite{eitel2015multimodal}.

Furthermore, it is demonstrated that the presented work can be extended to predict depth on top view zoomed images that can be utilized for a better understanding of surroundings when viewed from cranes. Here stereo zoom camera is very challenging to use as it requires careful calibration of cameras depending on the zoom, and Lidar, as well as normal stereo cameras, can not be used due to the distance. However, safety is of utmost concern, due to the sheer size of a crane and weight of the hoisted objects.

\begin{figure*}[tp]% AFAICT, adding ! does nothing useful
 \vspace*{\dimexpr -0.2in-\topmargin-\headheight-\headsep}%
 \makebox[\textwidth]{%
   \includegraphics[width=0.7\paperwidth]{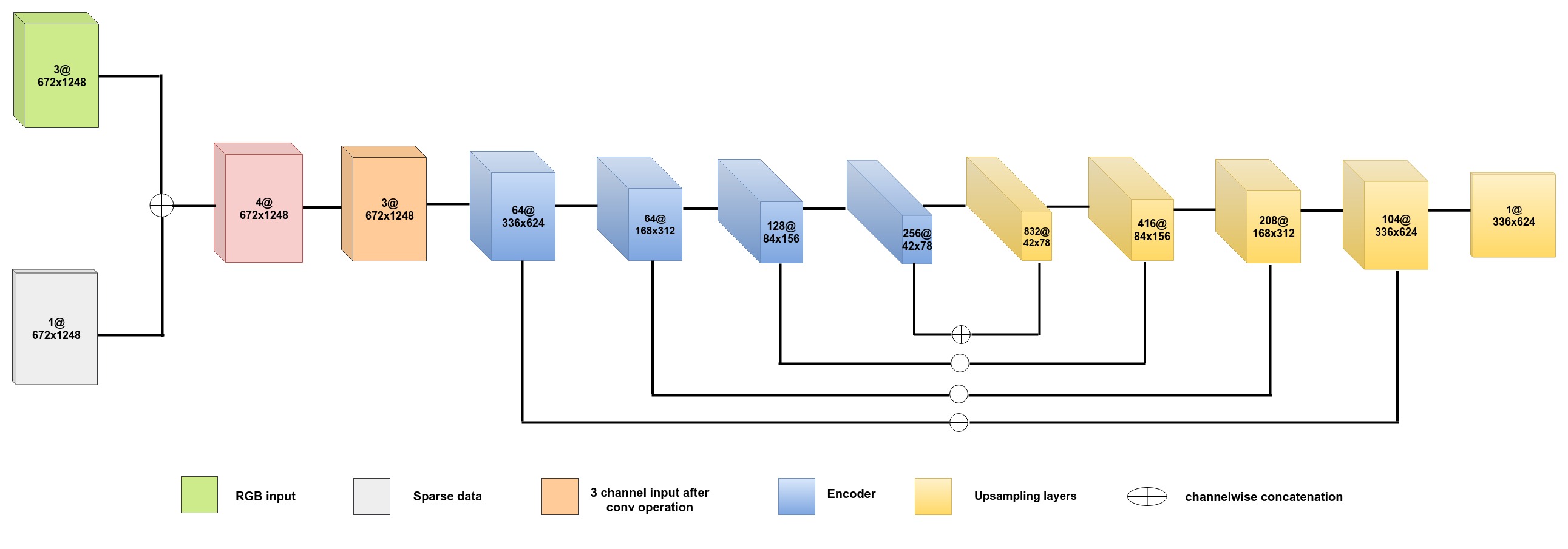}}
 \caption{Encoder-Decoder architecture with RGB and sparse data fusion as input. For
networks using only 3 channel input sparse data branch should be removed.}% do not put inside \makebox
 \label{Dense-Depth}
\end{figure*}

\section{RELATED WORK}
%Intro needed? First part of Monocular can be put here
Depth estimation and image reconstruction from a single RGB image is an "ill-posed problem" as many scenes in 3D planes represent the same 2D plane because of scale ambiguity. There have been some significant works of depth estimation using stereo and depth cameras but they have their limitations. In recent times, monocular depth estimation algorithms using Deep Learning has gained much recognition as they do not rely on hand-crafted features and these are mostly extracted using a Convolutional Neural Network resulting in superior results.\\
\textbf{Monocular Depth Estimation} - One of the early work was proposed by Saxena et al. \cite{NIPS2005_17d8da81} involving the use of probabilistic models such as Markov Random Field (MRF) or Conditional Random Field (CRF) on handcraft features extracted from a single camera image.
Eigen et al.\cite{EigenF14} proposed the two-stack convolutional neural network and directly regressed the pixels for depth estimation tasks.
Laina et al. \cite{LainaRBTN16} developed a deep residual network using ResNet as an encoder and a custom decoder. They proposed an "efficient scheme for upconvolutions using skip connections to create up-projection blocks for the effective upsampling of feature maps".
Fu et al. \cite{DBLP:journals/corr/abs-1806-02446} modified the depth learning problem from the regression problem to the ordinal quantized regression problem. Supervised methods for monocular depth estimation require the creation of groundtruth for training which is an expensive and time-consuming task. To overcome this problem, many unsupervised and self-supervised learning approaches for depth estimation have been proposed which do not require ground truth depth at training time  \cite{Goddard,GargBR16}. But it requires a stereo image pair for training, which is computationally expensive, and using images of the same modality are prone to errors in noisy environments.
\\
%However, only using one modulitiy is prominent to errors if the images are bad e.g. due to bad weather.
\iffalse
\textbf{Unsupervised Depth Estimation} - 
%Why is unsupervised so much better -> problem with GT for depth estimation. Usuayll from Lidar. Maybe fuse this directly with Monocular.
Recently . Garg et al. \cite{GargBR16} proposed
an unsupervised approach for depth estimation by training a deep convolutional network
similar to autoencoder, where stereo images pairs act as input and target. They trained
by generating an inverse warp of the target image by moving pixels from the right image along the scan-line
to match the source input through reconstruction loss. Goddard et al. \cite{Goddard} introduced a
novel method to predict disparities for both left and right images by exploiting epipolar geometry constraints when trained with an image reconstruction loss. \\
\fi
\textbf{Multi-Modal Depth Estimation} - Some deep learning models, when trained with data from different modality perform exceptionally well for tasks such as semantic segmentation and depth estimation utilizing vital information from both the modalities. For depth estimation tasks, sparse depth input from Lidar's or Radar's could be propitious when fused with noisy RGB images. Ma et al. \cite{Ma2017SparseToDense} proposed a supervised learning approach to train an encoder-decoder model by fusing sparse data with RGB sampled randomly from the ground truth depth image. Following their previous work, they trained a network \cite{ma2018self} with a semi-dense groundtruth depth map with a fusion of sparse Lidar point cloud and RGB image as inputs.
In comparison with the previous works, the proposed method predicts a full-resolution depth image, explores the importance of sparse measurements of Radar when fused with camera images that could supplement RGB images in areas with low visibility by providing distance data. Thus, helping in achieving low-level autonomy in the vehicles. 
\iffalse
The presented work also focuses on the depth estimation on top view images. Heavy construction vehicles such as cranes usually used for lifting heavy loads can reach up to 100 meters above the ground. Zoom-cameras are vital for such vehicles for a better view of the surroundings when the crane booms move ups and downs for lifting and moving heavy loads from one place to another. An efficient model that could predict correct depth information using zoomed images could be highly beneficial for autonomous and manual cranes to have a better understanding of the environment for preventing fatal accidents.
\fi
%The last part of the related work is more in the light of the introdution. Put this maybe there, just reference it here.

\section{METHODOLOGY}
In this section, the proposed architecture will be described. Convolutional Neural Networks are used for dense depth prediction task. Different data augmentation techniques and loss functions used for efficiently training the network is also discussed.  

\subsection{Architecture}

In computer vision, dense problems that require per pixel predictions such as depth estimation and semantic segmentation are usually resolved by using encoder-decoder architecture. Encoder maps raw inputs into feature representation and downsamples the spatial resolution of the inputs. In contrast, the decoder takes those feature representations as inputs, then processes it and gives output that is the closest match of the original input. The working of the decoder is opposite to that of encoders. Decoders consist of one or
more layers that increase the spatial resolution of the downsampled feature representations from the encoder. The final structure is based on the work done by Alhashim et al. \cite{Alhashim2018} as it utilizes the transfer learning approach and achieved state-of-the-art accuracy in RGB-based depth prediction for the NYU dataset. The network is modified so that it could fit the presented problem with fused input of different modalities. The encoder used in this work is DenseNet-169 pre-trained on Imagenet \cite{deng2009imagenet}, although experiments with ResNet-50 were also conducted but the errors were comparatively high in comparison to the model trained on DenseNet. As the pre-trained networks are restricted to 3 channel inputs, a 1x1 convolution layer has been used to truncate four-channel input to three-channel before passing the input to the network. A 1x1 filter only has a single parameter for each channel of the input, so it is widely considered for dimensionality reduction. Separable convolutions were attempted instead of standard convolution layer but it resulted in the substandard performance of the network. The resultant feature map extracted from the encoder is passed through series of upsampling blocks, comprising of a bilinear upsampling layer, followed by two convolution layers of 3x3 filters with output channels set to half the channels of the input and the result from the upsampling layer is concatenated with the output of the encoder having same spatial dimension similar to \cite{RFB15a}. A 3x3 convolution filter is favored as it can look few pixels at a time extracting the vast amount of local features that are highly useful in later layers. Fig. \ref{Dense-Depth} gives an overview of the architecture.

\subsection{Data Augmentation}

Data Augmentation is a strategy of increasing the amount and induce diversity in the
datasets for training Deep Learning models. For this work, augmentation is done in an online manner with transformations, such as random horizontal flipping, random contrast, and brightness in the range of [0.9, 1.1] and [0.75, 1.25] respectively, are used with fifty percent chance.

\subsection{Loss Functions}
Neural Networks are trained by minimizing the difference between the groundtruth and the predictions from the network.
A standard loss function for regression tasks %considers the difference between ground truth depth map and prediction of the depth map from the network.%
are $L_1$ and $L_2$ loss. Different variants of loss functions such as scale-invariant loss \cite{EigenPF14}, inverse Huber loss \cite{zwald2012berhu}, the combination of smoothness and reconstruction loss \cite{Goddard} can be found in depth estimation literature.
For our work, we have used loss functions similar to \cite{Alhashim2018} that consider penalizing the distortions at the boundaries of the objects on the image while focusing on minimizing the error between the groundtruth and predictions generated from the network. 
For training our network we used a composition of three different loss functions. 
If $d^*$ is the predicted depth and $d$ is the groundtruth depth, then the loss function is defined as
\begin{equation}
\hspace{-0.25 cm}
\label{eq:test}
L_{total}(d, d^{*})=L_{ssim}(d, d^{*})+L_{edge}(d, d^{*})+L_{pixel}(d, d^{*}))
\end{equation}
The first loss term is Structural Similarity Loss $(L_{ssim})$. $L_{ssim}$ has been used commonly for
image reconstruction tasks, which is based on the Structural Similarity (SSIM) index used for measuring the similarity between two images. The SSIM index can be viewed as a
quality measure for comparison between the two images provided \cite{10.1109/TIP.2003.819861}.
The second loss term $(L_{edge})$ is defined over the image gradient of the predictions and the depth groundtruth. It is basically the directional change in the intensity of the image \cite{wikiImage_gradient}. Often there is a discontinuity at the boundary of the object, so it is necessary to penalize the errors around the edges more.
For the third loss term $(L_{pixel})$, three different losses ${L}_{1}$, ${L}_{2}$, and Berhu loss \cite{zwald2012berhu} has been examined.
The difference between the ${L}_{1}$ and ${L}_{2}$ loss functions is that ${L}_{2}$ penalizes larger errors, but is more tolerant of small errors, whereas ${L}_{1}$ loss
does not produce a high error for large predictions error. ${L}_{1}$ loss tends to make the reconstructed image less blurry as compared to ${L}_{2}$ loss.

\iffalse
\begin{equation}
L_{1}(d, d^{*})=\frac{1}{n} \sum_{p}^{n}\left|d_{p}-d^{*}_{p}\right|
\end{equation}

\begin{equation}
L_{2}(d, d^{*})=\frac{1}{n} \sum_{p}^{n}\left|d_{p}-d^{*}_{p}\right|
\end{equation}
\fi
The Berhu loss or reverse Huber loss is a combination of ${L}_{1}$  and ${L}_{2}$ loss as described in the equation \ref{eq:berhu}. It acts as a ${L}_{1}$  loss when the difference between groundtruth and predictions falls below the threshold, otherwise acts as ${L}_{2}$  when the error exceeds that threshold.

\begin{equation}
\label{eq:berhu}
\mathcal{B}(x)=\left\{\begin{array}{ll}
|x| & |x| \leq c \\
\frac{x^{2}+c^{2}}{2 c} & |x|>c
\end{array}\right.
\end{equation}
 
The network is trained using reciprocal depth to compensate the problem of large errors resulting from bigger groundtruth depth values \cite{Alhashim2018} . The
final depth value is obtained by taking the inverse of the obtained prediction. 
\iffalse
The target depth map is defined as d = h/d* where d* is the original groundtruth, and h is the hyperparameter, which has been initialized as 10 for the synthetic dataset and 50 for the
real dataset.
\fi

\section{Experiments}
\subsection{Datasets}

\textbf{Nuscenes} is a publicly available large scale dataset for autonomous driving by Aptiv Autonomous Mobility \cite{nuscenes2019}. The dataset is collected from Boston and Singapore, two cities are known for their dense traffic situations and highly arduous driving conditions. Nuscenes is the only real recorded dataset available publicly that provides Radar data. Ground-truth depth maps have been generated by Lidar reprojection and missing values of depth are filled using the colorization method \cite{journals/tog/LevinLW04}. 

\textbf{KITTI} is one of the benchmark autonomous driving datasets available for depth completion and estimation, 3D object detection, and scene flow estimation tasks \cite{Geiger2012CVPR}. One of the significant advantages of using the Kitti dataset for depth estimation is the availability of semi-dense depth map groundtruth, which is obtained by registering 11 consecutive Lidar point cloud using Iterative
Closest Point Algorithm.  Although this work focuses on incorporating Radar data with RGB images for depth estimation, the impact of the Lidar cloud point in enhancing the accuracy of depth estimation using the same network when fused with RGB images is also examined. 
\\
\textbf{Synthetic Dataset}  has been created through the Carla simulator with nearly perfect groundtruth depths. Carla is an open-source emulator built on
top of the Unreal Engine 4 (UE4) gaming engine for autonomous driving research \cite{Dosovitskiy17}. With the recent release of CARLA in December 2019, it has added a low fidelity Radar sensor to complete its sensor suites. The data is recorded on multiple scenes with different field of view, thus providing a versatile range of images recorded on adverse weather conditions, its corresponding depth map, and Radar distance data. Each Radar point cloud is projected into the image and the depth information from Radar measurements is described as an image-like representation. An unseen test set is created for evaluation, comprising of 600 images recorded on scenes with different weather conditions such as day, night, fog, and rain.

\textbf{Top-View Dataset}
is generated with a gaming engine, Unreal Engine 4 \cite{unrealengine}. The images are captured from the cameras located at the boom of the crane on a construction site. The role of the crane is to lower the boom, pick up the load, and move the load to the desired position. The camera takes a snapshot of the load and the ground while the boom is moving \cite{10.22260/ISARC2018/0123}.

\begin{table}[h!]
\begin{tabular}{lllll}
\textbf{Name}      & \textbf{Type}       & \begin{tabular}[c]{@{}l@{}}\textbf{Sensing}\\ \textbf{Modalities}\end{tabular}   & \textbf{GT}                                                                      & \textbf{Size}                                                  \\ \hline \hline
Nuscenes  & Real       & \begin{tabular}[c]{@{}l@{}}Camera, Lidar,\\ Radar\end{tabular} & \begin{tabular}[c]{@{}l@{}}Created using \\ Lidar frames\end{tabular} & \begin{tabular}[c]{@{}l@{}}3400\\ frames\end{tabular} \\ \hdashline 
Synthetic & Simulation & Camera, Radar                                                  & \begin{tabular}[c]{@{}l@{}}Created using\\ Carla\end{tabular}          & \begin{tabular}[c]{@{}l@{}}8K\\ frames\end{tabular}   \\ \hdashline 
KITTI     & Real       & Camera, Lidar                                                  & \begin{tabular}[c]{@{}l@{}}Created using \\ Lidar frames\end{tabular} & \begin{tabular}[c]{@{}l@{}}9K\\ frames\end{tabular}   \\ \hdashline 
Top-View  & Simulation & Camera                                                         & \begin{tabular}[c]{@{}l@{}}Created from\\ UE4\end{tabular}            & \begin{tabular}[c]{@{}l@{}}1200\\ frames\end{tabular} \\ \hline
\end{tabular}
\caption{ Overview of the datasets used in this work}
\end{table}

\subsection{Implementation Details}
The network is implemented using TensorFlow 2.0 \cite{DBLP:journals/corr/AbadiABBCCCDDDG16}.
and trained on the above-mentioned datasets on a single machine with 12GB NVIDIA TITAN X GPU and 32GB internal memory, to have a fair comparison of all methods. The weights for the encoder is initialized with  pre-trained weights from the model trained on the ImageNet dataset \cite{deng2009imagenet}.
In this work, the batch size of 2 is used throughout all the experiments. The networks are trained for 20 epochs for all the datasets except Top-View dataset which is trained for 30 epochs using Adam Optimizer \cite{Kingma2014AdamAM} with exponential decay rate for the first moment and
the second moment estimates $\beta_1$ = 0.9, $\beta_2$ = 0.999. A constant learning rate of 0.0001 was experimented for training
the network but the best performing results were achieved by reducing the learning rate to
20 percent after every 7 epochs. So, for all the networks used in this work, the learning rate of 0.0001 which is reduced to 20 percent of its previous value at the eighth epoch is used for training.

\subsection{Evaluation Metrics}
To evaluate the quantitative performance of the networks, standard metrics for depth
estimation are used as mentioned in various depth estimation literature \cite{EigenF14}. Each
of the used metrics is explained below where $d_i$ is the original depth of the pixel in depth
image d and $d_{i}^{*}$ is the pixel of the predicted depth $d^*$ and N is the total number of pixels for each depth image.

\begin{itemize}
    \item Absolute Relative Distance (ARD): $\frac{1}{N} \sum_{1=1}^{N} \frac{\left|d_{i}-d_{i}^{*}\right|}{d_{i}^{*}}$

\item Squared Relative Difference (SRD):$ \frac{1}{N} \sum_{1=1}^{N} \frac{\left|d_{i}-d_{i}^{*}\right|^{2}}{d_{i}^{*}}$

\item Root Mean Square Error (RMSE): $\sqrt{\frac{1}{N} \sum_{1=1}^{N}\left|d_{i}-d_{i}^{*}\right|^{2}}$

\item Threshold Accuracy: $\%  \text { of } d_{i} \text { s.t. }\\
\max \left(\frac{d_{i}}{d_{i}^{*}}, \frac{d_{i}^{*}}{d_{i}}\right)<t h r, \text { where } \operatorname{thr} \in\left\{1.25,1.25^{2}, 1.25^{3}\right\}$ 

\end{itemize}

\section{RESULTS}
This section consists of empirical evaluations for all the mentioned datasets and discussion about the results of all the experiments performed using the described model. The experiments are supported by their qualitative and quantitative results.

\subsection{Architecture Evaluation}
\textbf{a). Evaluation on the Top-View dataset}

\begin{figure}[hbt!]
\hspace{-0.3 cm}
\centering
 \includegraphics[width=0.49\textwidth]{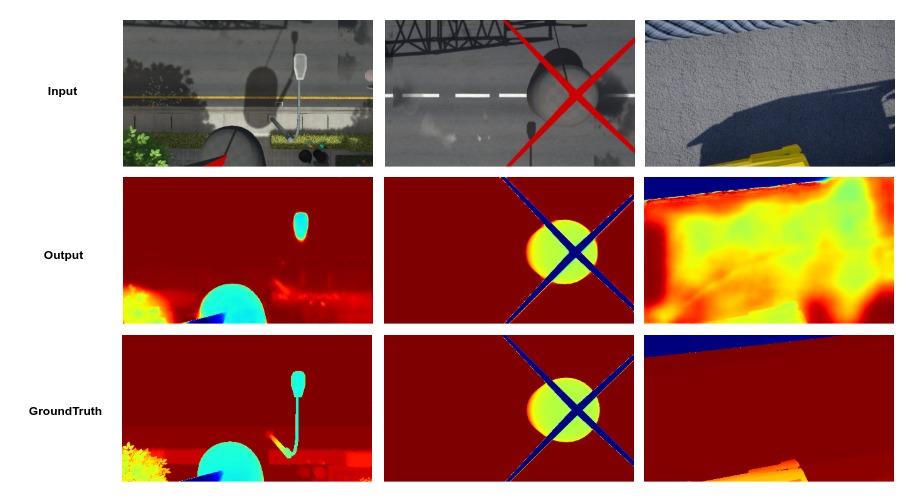}
  \caption{  Predictions on Top-View dataset.
 From top to bottom: (Input) - RGB images recorded from the camera located at the boom of the crane. (Output) - RGB based predictions. (Groundtruth) - Near perfect depth map generated from Unreal Engine. Note that, colormap is used for visualization of the depth map and it does not encode original depth information of the object from the camera.}
  \label{top_view-results}
\end{figure}

In this section, the quantitative evaluations on the Top-View dataset when trained with different loss functions are discussed. The network is trained with only RGB modality.
From the results, it can be inferred that the model trained with Berhu loss slightly outperformed the models trained with other loss functions in terms of RMSE, ARD, and SRD as listed in \ref{top-view}
. Predictions are depicted in Fig. \ref{top_view-results}
\begin{table}[hbt!]
\hspace{0.3 cm}
\begin{tabular}{llllllll}
\textbf{Modality}   & \hspace{-0.3 cm} \textbf{Loss} & \hspace{-0.3 cm} \textbf{RMSE} &\hspace{-0.3 cm} \textbf{ARD} &\hspace{-0.2 cm}\textbf {SRD}& \hspace{-0.1 cm} $\mathbf{\delta_1}$ &  $ \hspace{-0.1 cm} \mathbf{\delta_2}$ & $ \hspace{-0.1 cm} \mathbf{\delta_3}$ \\
\hline \hline
RGB        & \hspace{-0.3 cm} $L_1$   &\hspace{-0.3 cm} 0.257      & \hspace{-0.3 cm} 0.012     & \hspace{-0.3 cm} 0.017     & \hspace{-0.3 cm} 0.989   & \hspace{-0.3 cm} \textbf {0.998}    & \hspace{-0.3 cm} \textbf {0.999}                \\

RGB      &\hspace{-0.3 cm} $L_2$   & \hspace{-0.3 cm} 0.328    &\hspace{-0.3 cm} 0.017     &\hspace{-0.3 cm} 0.021    & \hspace{-0.3 cm} 0.986  & \hspace{-0.3 cm} {0.997}   & \hspace{-0.3 cm} {0.998}                          \\
RGB        &\hspace{-0.3 cm}Berhu   &\hspace{-0.3 cm} \textbf{0.252}     &\hspace{-0.3 cm} \textbf{0.011}     & \hspace{-0.3 cm} \textbf{0.015}     &\hspace{-0.3 cm} \textbf{0.992}   & \hspace{-0.3 cm} \textbf{0.998}  & \hspace{-0.3 cm} \textbf{0.999}
\\

\hline

\end{tabular}
\caption{ Quantitative performance of the network trained on Top-View dataset with different loss functions. Note that, L1, L2 and Berhu loss corresponds to the combination of that loss with SSIM and image gradient.
For every metric except threshold accuracies, lower values indicate better performance.}
\label{top-view}
\end{table}

\begin{figure}[hbt!]
\hspace{-0.4 cm}
\centering
  \includegraphics[width=0.5\textwidth]{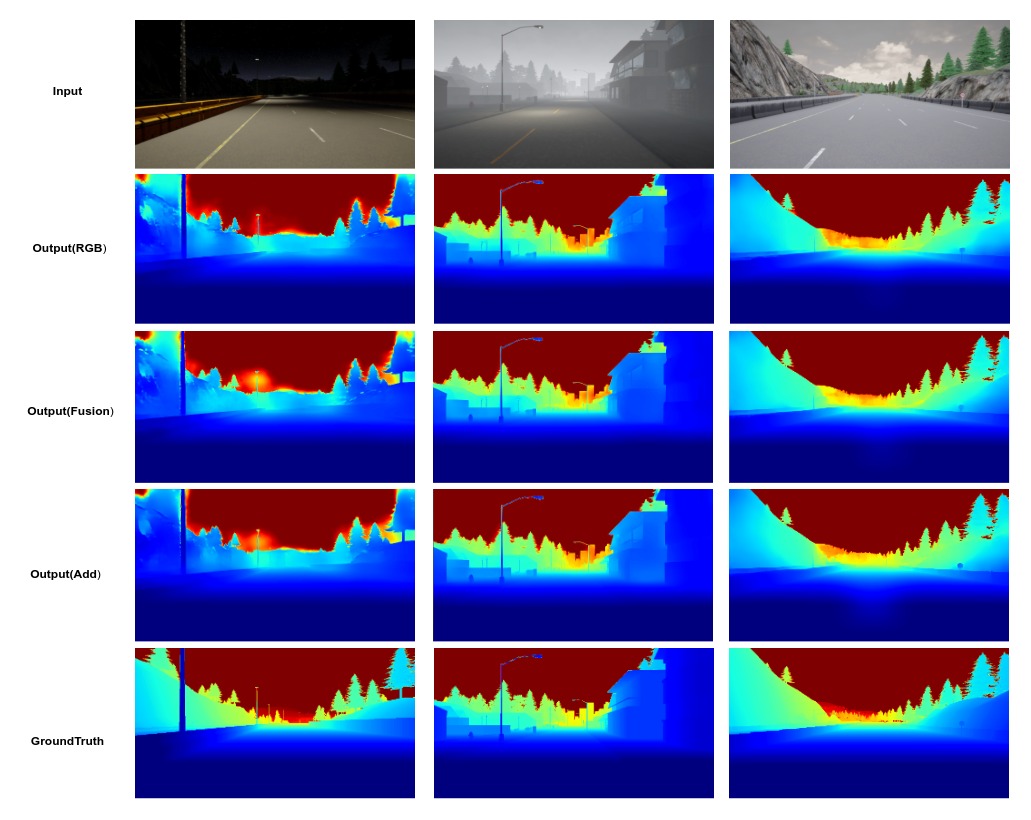}
  \caption{ Predictions on Synthetic dataset on scenes with different weather conditions.
 From top to bottom: (Input) - consisting of night, foggy and cloudy scenes. (Output(RGB)) - RGB based predictions. (Output(Fusion)) - Predictions on the fusion of RGB and sparse depth input. (Output(Add) - Predictions on element-wise addition of RGB and sparse depth input). (Groundtruth) - Near perfect depth map generated from Carla. Note that, colormap is used for visualization of the depth map and it does not encode original depth information of the object from the camera.}
  \label{Fig:synthetic}
\end{figure}

\textbf{b). Evaluation on the Synthetic dataset}
\\
In this section,  quantitative evaluations on the Synthetic dataset when trained with different loss functions are discussed. The fusion modality trained with $L_1$ loss is the best performing model in terms of ARD and threshold accuracies as compared to the model trained with other loss functions as mentioned in the Table \ref{tab:synthetic}. The modalities trained with Berhu loss also performs at par with modality trained on $L_1$ loss, the reason may be that the Berhu loss takes advantage of $L_1$ loss for very small errors and use $L_2$ otherwise. For Deep Learning models to be utilized on real-life applications, it is important to have a smaller inference time. The inference time for each frame on the model trained on RGB modality is 0.10 seconds and for the fusion modality, it is 0.11 seconds.
Qualitative results are depicted on Fig. \ref{Fig:synthetic} 

\begin{table}[hbt!]
\hspace{0.3 cm}
\begin{tabular}{llllllll}
\textbf{Modality}   & \hspace{-0.3 cm} \textbf{Loss} & \hspace{-0.3 cm} \textbf{RMSE} & \hspace{-0.3 cm}  \textbf{ARD} & \hspace{-0.2 cm}\textbf{SRD} & \hspace{-0.3 cm} $\mathbf{\delta_1}$ & \hspace{-0.2 cm}  $\mathbf{\delta_2}$ & \hspace{-0.1 cm} $\mathbf{\delta_3}$ \\
\hline \hline
RGB        & \hspace{-0.3 cm}  $L_1$   &\hspace{-0.3 cm}  0.715      &\hspace{-0.3 cm} 0.098     &\hspace{-0.3 cm} 0.126     &\hspace{-0.3 cm} 0.850   &\hspace{-0.3 cm} 0.945    &\hspace{-0.3 cm} 0.982                \\

RGB-Fusion &\hspace{-0.3 cm} $L_1$   &\hspace{-0.3 cm} 0.702    &\hspace{-0.3 cm} \textbf{0.093}     &\hspace{-0.3 cm} 0.120    &\hspace{-0.3 cm} \textbf{0.857}  &\hspace{-0.3 cm} \textbf{0.956}   &\hspace{-0.3 cm} \textbf{0.984}                          \\
RGB-Add    &\hspace{-0.3 cm} $L_1$   &\hspace{-0.3 cm} 0.723     &\hspace{-0.3 cm} 0.099     &\hspace{-0.3 cm} 0.129     &\hspace{-0.3 cm} 0.848   &\hspace{-0.3 cm} 0.943  &\hspace{-0.3 cm} 0.982
\\
\hdashline
\hdashline

RGB        &\hspace{-0.3 cm} $L_2$   &\hspace{-0.3 cm} 0.800      &\hspace{-0.3 cm} 0.106     &\hspace{-0.3 cm} 0.146     &\hspace{-0.3 cm} 0.825   &\hspace{-0.3 cm} 0.941    &\hspace{-0.3 cm} 0.982               \\
RGB-Fusion &\hspace{-0.3 cm} $L_2$   &\hspace{-0.3 cm} 0.796    &\hspace{-0.3 cm} 0.110     &\hspace{-0.3 cm} 0.142    &\hspace{-0.3 cm} 0.823  &\hspace{-0.3 cm} 0.954   &\hspace{-0.3 cm} 0.983                       \\
RGB-Add    &\hspace{-0.3 cm} $L_2$   &\hspace{-0.3 cm} 0.782     &\hspace{-0.3 cm} 0.105     &\hspace{-0.3 cm} 0.140     &\hspace{-0.3 cm} 0.830   &\hspace{-0.3 cm} 0.944  &\hspace{-0.3 cm} 0.983
\\
\hdashline
\hdashline
RGB        &\hspace{-0.3 cm}Berhu   &\hspace{-0.3 cm} 0.706      &\hspace{-0.3 cm} 0.096     &\hspace{-0.3 cm} 0.122     &\hspace{-0.3 cm} 0.855   &\hspace{-0.3 cm} 0.95    &\hspace{-0.3 cm} 0.983               \\
RGB-Fusion &\hspace{-0.3 cm}Berhu   &\hspace{-0.3 cm} 0.728     &\hspace{-0.3 cm} 0.101     &\hspace{-0.3 cm} 0.131    &\hspace{-0.3 cm} 0.84   &\hspace{-0.3 cm} 0.943  &\hspace{-0.3 cm} 0.983                    \\
RGB-Add    &\hspace{-0.3 cm}Berhu   &\hspace{-0.3 cm} \textbf{0.700}     &\hspace{-0.3 cm} 0.095     & \hspace{-0.3 cm} \textbf{0.119}     &\hspace{-0.3 cm} 0.856   &\hspace{-0.3 cm} 0.952  &\hspace{-0.3 cm} 0.983
\\

\hline

\end{tabular}
\caption{ Quantitative performance of the network trained on Synthetic dataset with different loss functions. RGB modality refers to the model trained only on camera images, RGB-Fusion modality refers to channel-wise concatenation of RGB and Radar data, whereas RGB-Add modality refers to the element-wise addition of RGB and Radar data. 
For every metric except threshold accuracies, lower values indicate better performance. Note that, $L_1$, $L_2$ and Berhu loss corresponds to the combination of that loss with SSIM and image gradient.}
\label{tab:synthetic}
\end{table}

%\newpage
\textbf{c). Evaluation on the KITTI dataset}

In this section, the quantitative evaluations on the KITTI dataset when trained on different modalities are presented and compared to the results with other benchmarked networks. The proposed network is evaluated on 652 images as specified by \cite{EigenF14}. During evaluation the predictions are upscaled from 192x624 to the original input resolution of 384x1248. The groundtruth used for evaluation is acquired by filling the missing depth values as mentioned in \cite{Alhashim2018}. The network is expected to perform much better when trained with the fusion of Lidar and RGB than the network trained on only RGB images because the input modality roughly is a subset of the groundtruth. The quantitative results are listed in \ref{kitti}.

\begin{table}[hbt!]
\hspace{0.1 cm}
\begin{tabular}{llllllll}
\textbf{Modality}   &\hspace{-0.2 cm} \textbf{Method}  & \hspace{-0.3 cm} \textbf{RMSE} &\hspace{-0.3 cm} \textbf{ARD} &\hspace{-0.2 cm} \textbf {SRD}& \hspace{-0.2 cm} $\mathbf{\delta_1}$ &  $ \hspace{-0.1 cm} \mathbf{\delta_2}$ & $ \hspace{-0.1 cm} \mathbf{\delta_3}$ \\
\hline \hline

RGB+sD &\hspace{-0.3 cm} Liao \cite{liao2016parse}   &\hspace{-0.3 cm} 4.50     & \hspace{-0.3 cm} {0.113}    & \hspace{-0.1 cm} -     & \hspace{-0.3 cm} {0.874}  & \hspace{-0.3 cm} {0.96}    & \hspace{-0.3 cm} {{0.984}}  \\

RGB+sD &\hspace{-0.3 cm} Ma \cite{Ma2017SparseToDense}   &\hspace{-0.3 cm} 3.378     & \hspace{-0.3 cm} \textbf{0.073}    & \hspace{-0.1 cm} -     & \hspace{-0.3 cm} {0.935}  & \hspace{-0.3 cm} {0.976}    & \hspace{-0.3 cm} \textbf{{0.989}}  \\

RGB+sD  &\hspace{-0.3 cm} Ours & \hspace{-0.3 cm} \textbf{1.81}    &\hspace{-0.3 cm} 0.076     &\hspace{-0.3 cm} 0.261    & \hspace{-0.3 cm} \textbf{0.937}  & \hspace{-0.3 cm} \textbf{0.981}   & \hspace{-0.3 cm} \textbf{0.989}                          \\

\hline

\end{tabular} %Maybe introduce a catchy name instead of "ours"
\caption{ Comparison with state-of-the-art on the KITTI dataset with input consisting of RGB image and sparse depth. Note that, the evaluation is performed on 652 images as mentioned in \cite{EigenF14}. The groundtruth used for evaluation is acquired by filling the missing depth values as mentioned in \cite{Alhashim2018}.  Ma et al. \cite{Ma2017SparseToDense} has used a random subset of 3200 images for the final evaluation and the groundtruth is acquired by projecting Velodyne LiDAR measurements onto the RGB images. For every metric except threshold accuracies, lower values indicate better performance.}
\label{kitti}
\end{table}

\begin{figure}[hbt!]
\hspace{-0.3 cm}
\centering
  \includegraphics[width=0.4\textwidth]{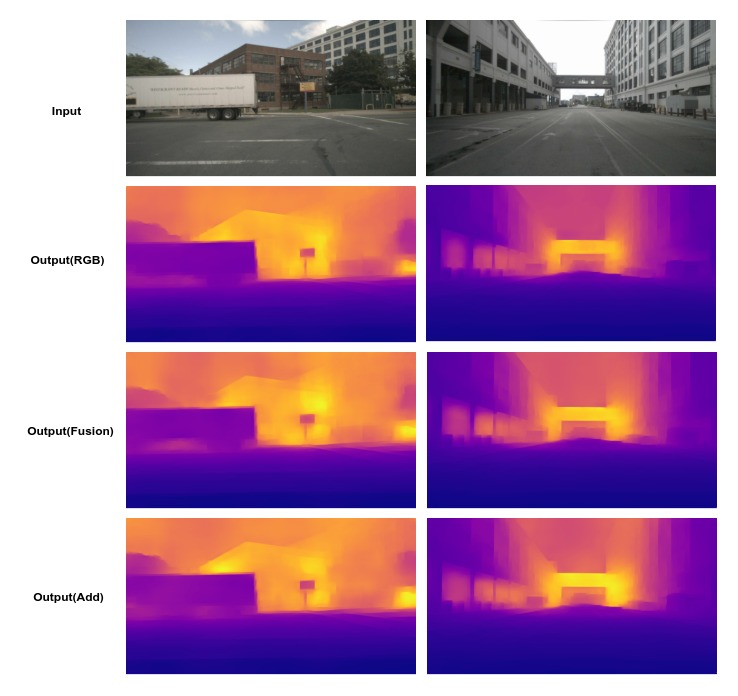}
  \caption{ Predictions on the Nuscenes dataset.
 From top to bottom: (Input) - RGB images. (Output(RGB)) - RGB based predictions. (Output(Fusion)) - Predictions on fusion of RGB and sparse depth input. (Output(Add) - Predictions on element-wise addition of RGB and sparse depth input).  Note that, colormap is used for visualization of the depth map and it does not encode original depth information of the object from the camera.}
  \label{Fig:nuscenes}
\end{figure}

\textbf{d). Evaluation on the Nuscenes dataset}
\\
In this section, the quantitative evaluations on the Nuscenes dataset when trained with different loss functions are presented. As groundtruth for Nuscenes is generated using Lidar point clouds and RGB images,
the modality of RGB is much more important as compared to Radar point clouds. In
comparison to all the modalities, the network trained with only RGB modality performed
better. The metric results for Radar fusion and addition modality are slightly substandard
as compared to a network trained only on RGB but on the visualization of predictions, the
results are much more comprehensible as compared to predictions from a network trained
only on RGB images and can be seen in Fig. \ref{Fig:nuscenes}

\begin{table}[hbt!]
\hspace{0.3 cm}
\begin{tabular}{llllllll}
\textbf{Modality}   &\hspace{-0.3 cm} \textbf{Loss} &\hspace{-0.3 cm} \textbf{RMSE} &\hspace{-0.3 cm} \textbf{ARD} &\hspace{-0.3 cm} \textbf{SRD}&\hspace{-0.3 cm} $\mathbf{\delta_1}$ &\hspace{-0.3 cm}  $\mathbf{\delta_2}$ &\hspace{-0.3 cm} $\mathbf{\delta_3}$ \\
\hline \hline
RGB        &\hspace{-0.3 cm} $L_1$   &\hspace{-0.3 cm} \textbf{2.27}      &\hspace{-0.3 cm} \textbf{0.089}     &\hspace{-0.3 cm} \textbf{0.309}     &\hspace{-0.3 cm} \textbf{0.903}   &\hspace{-0.3 cm} \textbf{0.985}    &\hspace{-0.3 cm} \textbf{0.997}                \\

RGB-Fusion &\hspace{-0.3 cm} $L_1$   &\hspace{-0.3 cm} 2.343    &\hspace{-0.3 cm} 0.093     &\hspace{-0.3 cm} 0.335    &\hspace{-0.3 cm} 0.895  &\hspace{-0.3 cm} 0.983   &\hspace{-0.3 cm} 0.996                          \\
RGB-Add    &\hspace{-0.3 cm} $L_1$   &\hspace{-0.3 cm} 2.273     &\hspace{-0.3 cm} 0.091     &\hspace{-0.3 cm} 0.317     &\hspace{-0.3 cm} \textbf{0.903}   &\hspace{-0.3 cm} \textbf{0.985}  &\hspace{-0.3 cm} 0.996
\\
\hdashline
\hdashline

RGB        &\hspace{-0.3 cm} $L_2$   &\hspace{-0.3 cm} 2.383      &\hspace{-0.3 cm} 0.094     &\hspace{-0.3 cm} 0.332     &\hspace{-0.3 cm} 0.891   &\hspace{-0.3 cm} 0.984    &\hspace{-0.3 cm} 0.996               \\
RGB-Fusion &\hspace{-0.3 cm} $L_2$   &\hspace{-0.3 cm} 2.41    &\hspace{-0.3 cm} 0.095     &\hspace{-0.3 cm} 0.344    &\hspace{-0.3 cm} 0.891  &\hspace{-0.3 cm} 0.982   &\hspace{-0.3 cm} 0.996                       \\
RGB-Add    &\hspace{-0.3 cm} $L_2$   &\hspace{-0.3 cm} 2.42     &\hspace{-0.3 cm} 0.094     &\hspace{-0.3 cm} 0.890     &\hspace{-0.3 cm} 0.848   &\hspace{-0.3 cm} 0.982  &\hspace{-0.3 cm} 0.996
\\

\hdashline \hdashline
RGB        &\hspace{-0.3 cm}Berhu   &\hspace{-0.3 cm} 2.356      &\hspace{-0.3 cm} 0.093     &\hspace{-0.3 cm} 0.339     &\hspace{-0.3 cm} 0.895   &\hspace{-0.3 cm} 0.981    &\hspace{-0.3 cm} 0.995               \\
RGB-Fusion &\hspace{-0.3 cm}Berhu   &\hspace{-0.3 cm} 2.416   &\hspace{-0.3 cm} 0.102     &\hspace{-0.3 cm} 0.374    &\hspace{-0.3 cm} 0.889  &\hspace{-0.3 cm} 0.981   &\hspace{-0.3 cm} 0.995                       \\
RGB-Add    &\hspace{-0.3 cm}Berhu   &\hspace{-0.3 cm} 2.372     &\hspace{-0.3 cm} 0.096     &\hspace{-0.3 cm} 0.353     &\hspace{-0.3 cm} 0.893   &\hspace{-0.3 cm} 0.981  &\hspace{-0.3 cm} 0.995
\\

\hline

\end{tabular}
\caption{ Quantitative performance of the network trained on the Nuscenes dataset with different loss functions. For every metric except threshold accuracies, lower values indicate better performance.}
\label{tab:nuscenes}
\end{table}

\section{CONCLUSIONS}
%Are the results now good enough for safety systems without stereo camera or lidar?
A convolutional neural network based depth prediction method for predicting dense depth images is proposed
by fusing RGB images with sparse depth input of Radar recorded on various weather conditions such as day, night, fog, cloudy, and rain. It was evaluated on the Nuscenes dataset, Synthetic dataset which is generated using CARLA and Top-View dataset comprising of images of construction sites generated from Unreal Engine. The experimental results reveal that by fusion of sparse input with RGB slightly enhanced the predictions of depth maps on the Synthetic dataset. Furthermore, we demonstrated that this model performs significantly better for predicting depth on the Top-View zoomed images captured from cranes, where stereo cameras or Lidars are not feasible to use. 
\iffalse
For the Nuscenes dataset, as the groundtruth is generated using Lidar point clouds with colorization of RGB, the modality of RGB plays a vital role when model trained with concatenation and addition of Radar point clouds with RGB images. It resulted in lower performance of the model trained on fusion modality in comparison to a model trained only on camera images. 
\fi
We believe that one avenue of future works is to examine the usage of neural architecture search as an encoder for dense depth prediction tasks. For the real-time operation of fusion methods, it is very essential to have a short latency for depth map predictions on embedded devices. Model compression by pruning methods could be explored to reduce the parameters of the network by masking the irrelevant features learned by the network. Moreover, self-supervised learning strategies on multi-modal datasets could be used as the creation of a groundtruth depth map is an expensive and tedious task.

\bibliographystyle{IEEEtran}
\bibliography{literature}

\end{document}